\documentclass[10pt,twocolumn,letterpaper]{article}

\usepackage{iccv}
\usepackage{float}
\usepackage{times}
\usepackage{epsfig}
\usepackage{graphicx}
\usepackage{amsmath}
\usepackage{amssymb}
\usepackage{multirow}
\usepackage{booktabs}

\usepackage[breaklinks=true,bookmarks=false]{hyperref}

\iccvfinalcopy 

\def\iccvPaperID{2542} 

\ificcvfinal\fi

\begin{document}

\title{BEV-DG: Cross-Modal Learning under Bird's-Eye View for Domain Generalization of 3D Semantic Segmentation}
\newcounter{cofn}\setcounter{cofn}{0}%
	\setcounter{footnote}{2}%
	\def\correspond{%
		\ifnum\value{cofn}=0%
		\footnote{Corresponding authors.}%
		\setcounter{cofn}{\value{footnote}}%
		\else%
		\footnotemark[\value{cofn}]%
		\fi%
	}%
\author{Miaoyu Li\textsuperscript{\rm 1},  
		Yachao Zhang\textsuperscript{\rm 2 $^\ddagger$},
		Xu Ma\textsuperscript{\rm 3},
		Yanyun Qu\textsuperscript{\rm 1 \thanks{Corresponding Author}},
		Yun Fu\textsuperscript{\rm 3} \\
		\textsuperscript{\rm 1}School of Informatics, Xiamen University \\
		\textsuperscript{\rm 2} Tsinghua Shenzhen International Graduate School, Tsinghua University\\
		\textsuperscript{\rm 3} Department of ECE, Northeastern University\\
		{\tt\small  limiaoyu@stu.xmu.edu.cn, yachaozhang@sz.tsinghua.edu.cn, yyqu@xmu.edu.cn}
	}

\maketitle
\ificcvfinal\thispagestyle{empty}\fi

\begin{abstract}
   Cross-modal Unsupervised Domain Adaptation (UDA) aims to exploit the complementarity of 2D-3D data to overcome the lack of annotation in a new domain. However, UDA methods rely on access to the target domain during training, meaning the trained model only works in a specific target domain. In light of this, we propose cross-modal learning under bird's-eye view for Domain Generalization (DG) of 3D semantic segmentation, called BEV-DG. DG is more challenging because the model cannot access the target domain during training, meaning it needs to rely on cross-modal learning to alleviate the domain gap. Since 3D semantic segmentation requires the classification of each point, existing cross-modal learning is directly conducted point-to-point, which is sensitive to the misalignment in projections between pixels and points. To this end, our approach aims to optimize domain-irrelevant representation modeling with the aid of cross-modal learning under bird's-eye view. We propose BEV-based Area-to-area Fusion (BAF) to conduct cross-modal learning under bird's-eye view, which has a higher fault tolerance for point-level misalignment. Furthermore, to model domain-irrelevant representations, we propose BEV-driven Domain Contrastive Learning (BDCL) with the help of cross-modal learning under bird's-eye view. We design three domain generalization settings based on three 3D datasets, and BEV-DG significantly outperforms state-of-the-art competitors with tremendous margins in all settings.
\end{abstract}

\section{Introduction}

Semantic segmentation of LiDAR point clouds is fundamental for numerous vision applications, such as robotics, autonomous driving and virtual reality. Given a LiDAR frame, the goal is to classify each point in the point cloud to produce semantic labels for them. 

\begin{figure}[t]
    \centering
\includegraphics[width=0.9\linewidth]{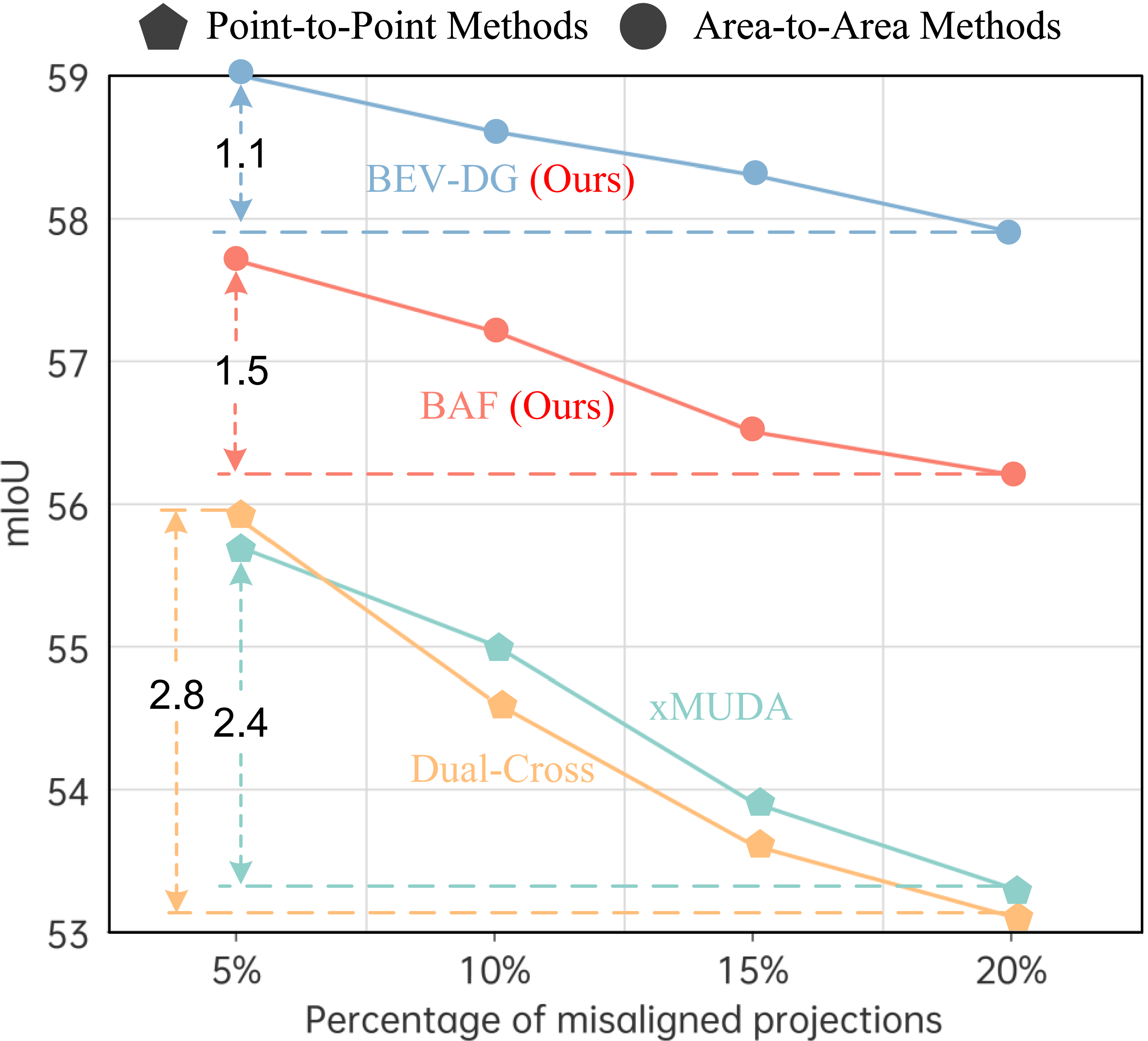}
    \caption{DG results of methods under different levels of point-to-pixel misalignment. The models are trained on A2D2 and SemanticKITTI datasets, and tested on nuScenes dataset. Area-to-area methods significantly outperform point-to-point methods under each level of misalignment. Moreover, point-to-point methods degrade more dramatically with the increasing misalignment.}
    \label{fig:disturb}
\end{figure}

Recently, some 3D semantic segmentation approaches have achieved impressive performance on several benchmark datasets \cite{  S3DIS,semantickitti,nuscenes,dai2017scannet,a2d2}. However, the training and test data of these methods are from the same dataset (domain). As each dataset has a different configuration of LiDAR sensors, these methods can significantly degrade under domain shift. Specifically, due to the number of laser beams varying from LiDAR to LiDAR, the density (resolution) of obtained point cloud is also quite different, which results in a large domain gap. To improve the generalization of the model, some Unsupervised Domain Adaptation (UDA) methods \cite{xmuda,dualcross,dscml,comlabel,sse} are proposed for 3D semantic segmentation in a single-modal or cross-modal manner. However, the training of these UDA methods relies on the target domain data, which makes them only generalize well to a specific target domain.

To this end, we are focused on investigating Domain Generalization (DG) for 3D semantic segmentation. Compared to UDA, DG is more challenging as it can not access the target domain during training, and the model should generalize well to an unseen target domain. Currently, many cross-modal UDA methods \cite{xmuda,dualcross,dscml,sse} are proposed for 3D semantic segmentation. To improve the robustness of the model to domain shift, they utilize cross-modal learning to prompt information interaction between two modalities (image and point cloud).
The mechanism behind this idea is that if one modality is sensitive to the domain shift while the other is robust, the robust modality can guide the sensitive modality. In light of this, we solve the DG problem using cross-modal learning on multi-modal data.

However, existing cross-modal learning is conducted point-to-point, using the projections between 2D pixels and 3D points to achieve cross-modal matching. Due to the inaccuracy of the extrinsic calibration between LiDAR and camera \cite{9196512}, there is more or less point-level misalignment in the projections. As a result, existing cross-modal UDA methods degrade significantly when extending them to DG task. Unlike UDA, which allows fine-tuning on the target domain, the target domain is unavailable in the DG setting. Thus these point-to-point cross-modal UDA methods are more sensitive to inaccurate cross-modal matching caused by point-level misalignment, as seen in Fig. \ref{fig:disturb}. Moreover, to model domain-irrelevant representations, some cross-modal UDA methods \cite{dscml,sse} introduce adversarial learning, which is well-known to be highly sensitive to hyperparameters and difficult to train.

To address these issues, we propose cross-modal learning under bird's-eye view for domain generalization of 3D semantic segmentation, which is inspired by 3D object detection methods \cite{pointpillars,3dcoco,polarnet} that use the additional bird's-eye view of one modality (point cloud) to better the target posture and boundary. For different modalities (image and point cloud), with the help of an auxiliary bird's-eye view, we alleviate the cross-modal matching error caused by point-to-pixel misalignment and optimize the domain-irrelevant representation modeling.
Specifically, we first propose BEV-based Area-to-area Fusion (BAF). Instead of conducting cross-modal learning point-to-point, we divide the point cloud and its corresponding image into areas with the help of a unified BEV space. And then, based on point-to-pixel projections, we match areas from two modalities to conduct area-to-area fusion. The cross-modal matching between areas has a higher fault tolerance for point-level misalignment. Because two projected point and pixel are more likely to be located in the same area than sharing the same accurate location. In this way, we significantly mitigate the influence of point-level misalignment and achieve accurate cross-modal learning in an area-to-area manner.

Furthermore, BEV-driven Domain Contrastive Learning (BDCL) is proposed to optimize domain-irrelevant representation modeling. First, with the aid of cross-modal learning under bird's-eye view, we generate the BEV feature map in a voxelized manner. This process is greatly affected by point cloud density, which makes the BEV feature map highly domain-relevant. Thus, using the BEV feature map to drive contrastive learning can provide stronger supervision for learning domain-irrelevant features. However, domain attribute, \textit{i.e.}, LiDAR configuration, is reflected in the global density of the point cloud. Therefore, we propose Density-maintained Vector Modeling (DVM) to transform the BEV feature map into a global vector that maintains density perception. Then, we build contrastive learning that constrains consistency between BEV vectors before and after changing domain attributes. Moreover, as the BEV vectors contain domain-retentive multi-modal information, BDCL can push both 2D and 3D networks to learn domain-irrelevant features jointly. 

Our contributions can be summarized as follows:

$\bullet$  We propose BEV-DG for domain generalization of 3D semantic segmentation. With the aid of cross-modal learning under bird's-eye view, we optimize domain-irrelevant representation modeling in a constraint manner. 

$\bullet$  To relive the cross-modal learning from the suffering of misalignment in point-to-pixel projections, we propose BEV-based area-to-area fusion. The accurate area-to-area cross-modal learning under bird's-eye view can more efficiently promote the information interaction between modalities to confront the domain shift.

$\bullet$  We propose BEV-driven domain contrastive learning, where the Density-maintained Vector Modeling (DVM) is introduced to generate the global vector that sufficiently embodies domain attributes. Furthermore, with the help of Density Transfer (DT), we build contrastive learning based on these vectors, pushing 2D and 3D networks to learn domain-irrelevant features jointly.

$\bullet$  We design three generalization settings based on three 3D datasets and provide the results of some competitors by extending cross-modal UDA methods to the DG setting. Extensive experimental results demonstrate that BEV-DG significantly outperforms the baseline and state-of-the-art competitors in all generalization settings.
\begin{figure*}[t]
\centering
\includegraphics[width=\linewidth]{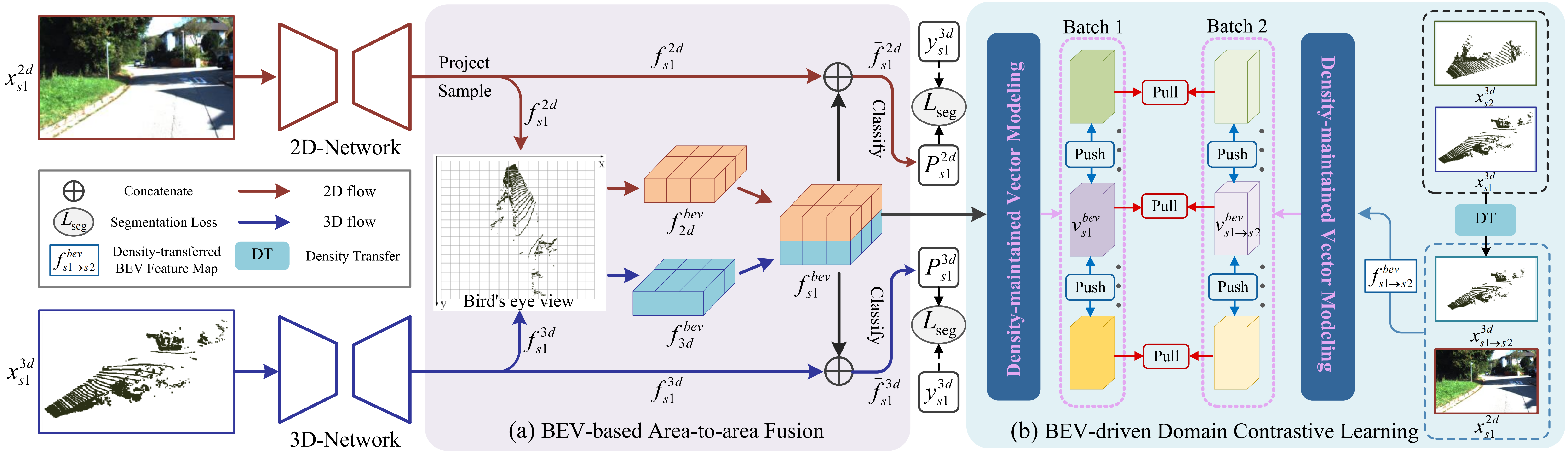}

\caption{The framework of BEV-DG. When it is trained on the first source domain, the input is paired samples $\mathbf{\left \{x_{s1}^{2d}, x_{s1}^{3d}\right \}}$ with 3D label $y_{s1}^{3d}$. Through module (a), the BEV feature map $f_{s1}^{bev}$ is obtained based on 2D and 3D features $\mathbf{\left \{f_{s1}^{2d}, f_{s1}^{3d}\right \}}$. Using point-to-pixel projections, we sample from the dense 2D feature map to generate 2D features of the length $N$, \textit{i.e.}, the number of 3D points. Then, $f_{s1}^{bev}$ is fused with $f_{s1}^{2d}$ and $f_{s1}^{3d}$, respectively, to generate predictions. Moreover, $f_{s1}^{bev}$ is fed into the module (b) to generate density-maintained BEV vector $v_{s1}^{bev}$ to drive contrastive learning for domain-irrelevant representation modeling. } 
\label{fig:framework}
\end{figure*}
\section{Related Work}
\noindent{\textbf{Point Cloud Semantic Segmentation. }}
Semantic segmentation of point clouds based on deep learning has made significant progress in recent years\cite{sparseconvnet,RandLA,landrieu2018large,qi2017pointnet++,shellnet}. However, these methods rely on fine-grained annotations, which are costly to obtain. To this end, weakly supervised semantic segmentation has attracted increasing attention \cite{shi2021label,wei2020multi,xu2020weakly, zhang, psd} as it reduces the reliance on labels. These fully or weakly supervised methods achieve impressive performance on current datasets. However, the training and test data of these methods are taken from the same dataset. Due to the difference in distribution between the datasets, whether the model trained on one dataset can generalize well to others is a critical question to consider.

\noindent{\textbf{Domain Adaptation for 3D Point Clouds. }}
The past few years have seen an increasing interest in Unsupervised Domain Adaptation (UDA) techniques for 3D vision tasks. 3D-CoCo \cite{3dcoco} utilizes highly transferable BEV features to extend the approach of contrastive instance alignment to point cloud detection, aiming to push the model to learn domain-irrelevant features. For point cloud segmentation, Complete and Label \cite{comlabel} designs a completion network to recover the underlying surfaces of sparse point clouds. The recovered 3D surfaces serve as a canonical domain, from which semantic labels can transfer across different LiDAR sensors. However, these UDA methods only utilize 3D modality (point cloud), neglecting the value of 2D images.

As 3D datasets often consist of 3D point clouds and corresponding 2D images, utilizing multi-modality to solve domain shift problems for point clouds is convenient. Some cross-modal UDA methods \cite{xmuda,dualcross,dscml,sse} have recently been proposed for 3D semantic segmentation. Under the hood of such methods lies the same spirit of cross-modal learning, \textit{i.e.}, the fusion of multi-modal information. With the help of projections between points and pixels, these methods achieve cross-modal learning by constraining the consistency between point and pixel predictions. This point-to-point manner is significantly affected by misalignment in projections. To this end, in this paper, we focus on achieving more accurate cross-modal learning that is less influenced by point-level misalignment.

\noindent{\textbf{Domain Generalization. }}
Domain Generalization (DG) considers the generalization ability of the model to unseen domains. Compared to UDA, the model can not access data from the target domain during training. The common methods include learning domain-irrelevant features based on multiple source domains \cite{hu2020domain,lehner20223d,li2018learning,li2018domain,motiian2017unified,piratla2020efficient} and enlarging the available data space with augmentation \cite{carlucci2019domain,qiao2020learning,shankar2018generalizing,zhou2020deep,zhou2020learning}. Recently, Some approaches also exploit regularization with meta-learning \cite{li2019episodic} and Invariant Risk Minimization (IRM) \cite{arjovsky2019invariant} framework for DG. However, these methods are focused on 2D vision tasks. Relatively little work has been done to study domain generalization for 3D point clouds. 3D-VField \cite{lehner20223d} attempts to improve the generalization of 3D object detectors to out-of-domain data by using proposed adversarial augmentation to deform point clouds during training. Compared to these methods, our BEV-DG aims to utilize multi-modal data to investigate DG for 3D semantic segmentation.

\section{Method}
\subsection{Overview of BEV-DG}

\noindent{\textbf{Problem Definition. }}
The problem assumes the presence of paired 2D images and 3D point clouds in Domain Generalization (DG) for 3D semantic segmentation. The DG task aims to exploit the knowledge from two source domains, $S1$ and $S2$, to generalize to the target domain $T$. 
All domains contain images and point clouds $\left \{x^{2d},x^{3d}\right \}$. For simplicity, we only use the front camera image and the LiDAR points that project into it. The model is trained on $S1$ and $S2$, where only the 3D label $y^{3d}_{s1/2}$ exists, and tested on $T$. In the DG task, the model can not access the target domain during training, \textit{i.e.}, $T$ is unseen. Due to images and point clouds being heterogeneous, $x^{2d}$ and $x^{3d}$ are fed into 2D network $h^{2d}$ and 3D network $h^{3d}$ to output segmentation results $P^{2d}$ and $P^{3d}$, respectively. Cross-modal DG aims to utilize the complementarity of $x^{2d}$ and $x^{3d}$ to improve both $P^{2d}$ and $P^{3d}$. Moreover, the projections between 2D pixels and 3D points are provided by data prior.

\noindent{\textbf{Method Overview. }}
 Our approach consists of BEV-based Area-to-area Fusion (BAF) and BEV-driven Domain Contrastive Learning (BDCL), aiming to optimize domain-irrelevant feature modeling with the aid of cross-modal learning under bird's-eye view. The overall framework is depicted in Fig. \ref{fig:framework}. Take the training on the first source domain as an example, and the training on the second domain is the same. Given the paired 2D and 3D samples (\textit{i.e.},  $x_{s1}^{2d}$ and $x_{s1}^{3d}$), we first input them into BAF to generate BEV feature map (\textit{i.e.}, $f_{s1}^{bev}$) and output segmentation results (\textit{i.e.},  $p_{s1}^{2d}$ and $p_{s1}^{3d}$). Specifically, we process 2D and 3D features (\textit{i.e.},  $f_{s1}^{2d}$ and $f_{s1}^{3d}$) in a unified BEV space to generate 2D and 3D BEV feature maps (\textit{i.e.},  $f_{2d}^{bev}$ and $f_{3d}^{bev}$). Next, we concatenate them to produce $f_{s1}^{bev}$ and fuse it with $f_{s1}^{2d}$ and $f_{s1}^{3d}$ respectively, obtaining  $\Bar{f}_{s1}^{2d}$ and $\Bar{f}_{s1}^{3d}$ to further generate the predictions. In addition, $f_{s1}^{bev}$ is fed into BDCL to help the networks learn domain-irrelevant features. Through Density-maintained Vector Modeling (DVM), we use generated BEV vector (\textit{i.e.}, $v_{s1}^{bev}$) to form positive and negative pairs with other BEV vectors. For the positive pair, we generate a density-transferred BEV vector (\textit{i.e.}, $v_{s1\to s2}^{bev}$) with the help of Density Transfer (DT), which transforms the density of $x_{s1}^{3d}$ into the second domain point cloud $x_{s2}^{3d}$.

\subsection{BEV-based Area-to-area Fusion}
    Previous methods conduct cross-modal learning in a point-to-point manner based on the projections between 3D points and 2D pixels. However, due to the inaccuracy of extrinsic calibration between LiDAR and camera \cite{9196512}, more or less point-level misalignment exists, hindering the effectiveness of such methods. In light of this, we propose BAF to conduct cross-modal learning under bird's-eye view in an area-to-area manner. In this way, we can effectively mitigate the influence of point-level misalignment and achieve more accurate cross-modal learning.

\noindent{\textbf{BEV Transformation. }}
Camera captures data in perspective view and LiDAR in 3D view. This view discrepancy makes it difficult to appropriately divide the areas of the image and point cloud and set the matching relationship between them. To this end, we introduce a unified BEV space to transform the image and point cloud into the same view.
For point cloud $x_{s1}^{3d}$, we first quantize it along the x-axis and y-axis to generate pillar voxels evenly, as shown in module (a) of Fig. \ref{fig:framework}. These voxels can be regarded as different areas of point cloud under the bird's-eye view. As a result, the points are assigned to these areas according to their coordinates. The feature of a voxel is obtained by max-pooling the features of points inside it. For example, the feature in the $i,j$-th grid cell is:
\begin{equation}
  \begin{aligned}
      f_{i,j}^{3d}=& MAX(\left \{ h^{3d}(p^{3d})\mid (i-1)w<p_x^{3d}<iw, \right.\\ 
      &\left.(j-1)w<p_{y}^{3d}<jw\right \}),
  \end{aligned}
  \label{eq:voxel feature}
\end{equation}
where $f_{i,j}^{3d}\in \mathcal{R} ^{1\times C_{3d}}$. $C_{3d}$ is the number of channels of 3D features. $MAX$ denotes the max pooling operation. The size of a grid cell is $w\times w$. $p_x^{3d}$/$p_y^{3d}$ is the x/y coordinate of 3D point $p^{3d}$, \textit{i.e.}, its locations in the BEV space. Finally, the BEV feature map of $x_{s1}^{3d}$ can be formulated as follows:
\begin{equation}
    f_{3d}^{bev}=\left \{ f_{i,j}^{3d}\mid i\in \left \{ 1,2\dots ,W \right \}, j\in \left \{ 1,2\dots ,L \right \}   \right \},
    \label{eq:BEV feature}
\end{equation}
where $f_{3d}^{bev}\in \mathcal{R}^{W\times L\times C_{3d}}$. $W$ and $L$ denote the number of grid cells along the x-axis and y-axis, respectively. 

How to transform the image into bird's-eye view is a challenging problem. To tackle it, existing methods \cite{bevformer,bevfusion} usually utilize depth estimation or transformer, which are very complex and costly. In contrast, we simply use the point-to-pixel projections provided by data prior to conduct view transformation for images. Specifically, for a pixel $p^{2d}$ in image $x_{s1}^{2d}$, which projects to point $p^{3d}$, its accurate locations in the BEV space may be different from $p_{x}^{3d}$ and $p_{y}^{3d}$ due to misalignment. However, to transform the image into bird's-eye view, we only need to determine the proximate locations of pixels, \textit{i.e.}, the voxels in which pixels are located. A pillar voxel covers much more space than a point, and even if $p^{2d}$ mismatches $p^{3d}$, they are still likely located in the same voxel. So we determine the voxels where pixels are located based on the corresponding 3D points, effectively mitigating the influence of misalignment.
Finally, we can obtain 2D BEV feature map $f_{2d}^{bev}$ as follows:
\begin{equation}
  \begin{aligned}
      f_{i,j}^{2d}=& MAX(\left \{ h^{2d}(p^{2d})\mid (i-1)w<p_x^{3d}<iw,\right.\\ 
      &(j-1)w<p_{y}^{3d}<jw \left.  \right \}),
  \end{aligned}
  \label{eq:2D voxel feature}
\end{equation}
\begin{equation}
    f_{2d}^{bev}=\left \{ f_{i,j}^{2d}\mid i\in \left \{ 1,2\dots ,W \right \}, j\in \left \{ 1,2\dots ,L \right \}   \right \}.
    \label{eq:2D BEV feature}
\end{equation}

\noindent{\textbf{Area-to-area Fusion. }}
After BEV transformation, we divide the image and point cloud into areas using the same criteria and obtain 2D and 3D features of these areas, \textit{i.e.}, $f_{2d}^{bev}$ and $f_{3d}^{bev}$. Compared to point-to-point cross-modal learning, our method does not need to match pixels and points based on projections that may be misaligned. Instead, we just need to match their areas. Compared with sharing the same accurate location in BEV space, two projected point and pixel are more likely to be located in the same voxel (area), which means matching between areas based on point-to-pixel projections has a higher fault tolerance for point-level misalignment. So we directly concatenate $f_{2d}^{bev}$ and $f_{3d}^{bev}$, followed by a linear layer with ReLU, to achieve area-to-area fusion:
\begin{equation}
    f_{s1}^{bev}=ReLU(FC_1(f_{2d}^{bev}\oplus f_{3d}^{bev})),
    \label{eq: map fusion}
\end{equation}
where $f_{s1}^{bev}$ is the fusion BEV feature map of $x_{s1}^{2d}$ and $x_{s1}^{3d}$. Next, we further fuse this area-level information with initial point-level features for final semantic segmentation:
\begin{equation}
    \bar{f} _{p^{3d}}=ReLU(FC_2(h^{3d}(p^{3d})\oplus f_{i,j}^{bev})),
\end{equation}
where $f_{i,j}^{bev}$ is the feature of voxel where point $p^{3d}$ is located, \textit{i.e.}, the $i,j$-th feature in $f_{s1}^{bev}$. $\bar{f}_{s1}^{3d}$ consists of all $N$ fused point features $\bar{f} _{p^{3d}}$. The process to obtain $\bar{f}_{s1}^{2d}$ is identical. This fusion provides bird's-eye-view multi-modal contextual information for each point (pixel) in a point-to-area manner. As only matching between the point (pixel) with the area where it is located, this manner is also less susceptible to point-level misalignment. In summary, each stage of BAF effectively mitigates the impact of misalignment, achieving more accurate cross-modal learning.

\subsection{BEV-driven Domain Contrastive Learning}

 Previous methods usually utilize adversarial learning to model domain-irrelevant representations, which are highly sensitive to hyperparameters and difficult to train. In light of this, we introduce BDCL, which conducts contrastive learning between different domains and samples with the help of cross-modal learning under bird's-eye view. Specifically, we promote the consistency between samples before and after changing the domain attributes, providing additional supervision for learning domain-irrelevant features. For contrastive learning, the stronger the domain relevance of the sample features, the stronger the supervision will be. Thus we choose BEV feature map $f_{s1}^{bev}$ generated by cross-modal learning under bird's-eye view to drive the contrastive learning. It is produced in a voxelized manner, which makes it highly related to the point cloud density. Concretely, as the size of a voxel is fixed, the number of points inside it highly depends on the density. Therefore, compared to the initial point-level features $f_{s1}^{2d}/f_{s1}^{3d}$, $f_{s1}^{bev}$ has stronger domain relevance. Moreover, as $f_{s1}^{bev}$ contains domain-retentive multi-modal information, BDCL can push both 2D and 3D networks to learn domain-irrelevant features jointly. 
 
 Our proposed BDCL consists of two components: (1) Density-maintained Vector Modeling (DVM); (2) building contrastive learning to help 2D and 3D networks jointly learn domain-irrelevant features.

\begin{figure}[t]
    \centering
    \includegraphics[width=0.9\linewidth]{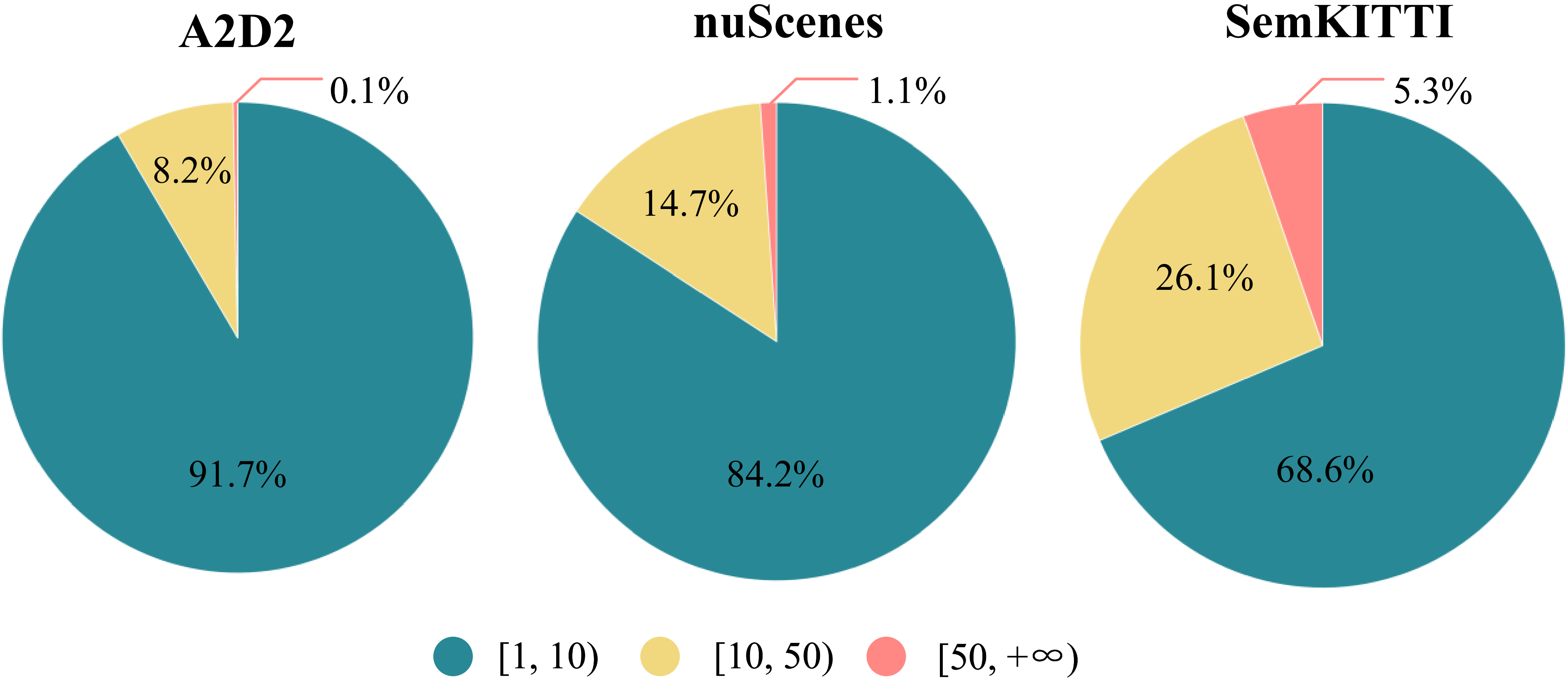}
    \caption{The distribution of areas in BEV space over dataset. We divide point clouds from three datasets into areas in the BEV space. Following that, we classify the areas into three groups according to the number of points inside them. Different distribution patterns can be seen clearly when we show the percentage of each type of area: points are spread more densely in nuScenes (32 beam) and SemanticKITTI (64 beam) than in A2D2 (16 beam). }
    \label{fig:pizza}
\end{figure}

\noindent{\textbf{Density-maintained Vector Modeling (DVM). }}
 As the LiDAR configuration determines the global density of the point cloud, the domain attribute of a sample should be characterized by its global feature. However, The BEV feature map consists of area-level features. Thus, we must transform it into a global vector that can embody domain attributes well. For a point cloud, the distribution of points inside it varies greatly. Concretely, the density of the part near the LiDAR is greater than the part away from the LiDAR. Thus, directly modeling the global vector from $f_{s1}^{bev}$ by equally treating points in different areas can not maintain the perception of density. In light of this, we propose DVM to transform $f_{s1}^{bev}\in \mathcal{R}^{W\times L\times C}$ into BEV vector $v_{s1}^{bev}\in \mathcal{R}^{1\times C}$ without undermining the density perception of feature. Specifically, in the BEV space, we find the density of the point cloud is well reflected in the distribution of areas, as shown in Fig. \ref{fig:pizza}. This distribution pattern helps us model density-maintained vector from BEV feature map $f_{s1}^{bev}$. More analysis of DVM can be seen in the supplementary material. We can generate the BEV vector as follows:
\begin{equation}
    \begin{aligned}
        v_{s1}^{bev} = &\frac{{{N_{[1,10)}}}}{{{N_{all}}}}MAX\left( {f_{[1,10)}^{bev}} \right) + \frac{{{N_{[10,50)}}}}{{{N_{all}}}}MAX\left( {f_{[10,50)}^{bev}} \right) \\ + &\frac{{{N_{[50,+\infty)}}}}{{{N_{all}}}}MAX\left( {f_{[50,+\infty)}^{bev}} \right),
    \end{aligned}
    \label{eq:vector}
\end{equation}
where $N_{[1,10)}$/$N_{[10,50)}$/$N_{[50,+\infty)}$ is the number of areas with $[1,10)/[10,50)/[50,+\infty)$ points inside. $N_{all}$ is the number of all areas. $f_{[1,10)}^{bev}$/$f_{[10,50)}^{bev}$/$f_{[50,+\infty)}^{bev}$ is the feature set of areas with $[1,10)/[10,50)/[50,+\infty)$ points inside.

\noindent{\textbf{Architecture of BDCL. }}
To form the negative and positive pairs of BEV vectors, we first utilize Density Transfer (DT) in Dual-Cross \cite{dualcross} to generate approximate BEV vectors of the other source domain. Concretely, we transform the densities of point clouds in the current batch into densities of point clouds from the other source domain, as depicted in module (b) of Fig. \ref{fig:framework}. In this process, the semantic content of point clouds and their corresponding images remain unchanged. Using these synthetic point clouds and their corresponding images, we can generate a new batch of density-transferred BEV vectors, which share the same semantic content but perceive the density of the other domain. On the one hand, we push the BEV vectors in the same batch (domain) away from each other in the representation space. Since a negative pair of samples come from a single domain and shares identical domain attribute, domain-irrelevant representations are learned to contrast them. On the other hand, we pull the BEV vectors that share the same semantic content but from different batch (domain) close in the representation space, promoting the networks to learn domain-irrelevant features jointly.
The contrastive loss of the first source domain can be formulated as follows:
\begin{equation}\label{conloss}
    \scriptsize
    \mathcal{L}_{ct}^{s1}=-\frac{1}{B}\sum_{i=1}^{B} log\frac{exp(v_{i}^{s1}\cdot v_{i}^{s1\to s2}/\tau )}{\sum\limits _{j=1}^{B}exp(v_i^{s1}\cdot v_j^{s1}/\tau )+\sum\limits _{k=1}^{B}exp(v_i^{s1\to s2}\cdot v_k^{s1\to s2}/\tau ) },
\end{equation}
where B is the batch size, $ v_i^{s1}$ is the $i$-th BEV vector in the first domain batch, and $v_i^{s1\to s2}$ is the corresponding density-transferred BEV vector of $v_i^{s1}$. $\tau$ is the temperature hyperparameter that controls the concentration level. The contrastive loss of the second source domain is the same.
\subsection{Overall Objective Function}
The segmentation loss of the first source domain can be formulated as follows:
\begin{equation}
    \mathcal{L}_{seg}^{s1}(x_{s1},y_{s1}^{3d})=-\frac{1}{NC} \sum_{n=1}^{N} \sum_{c=1}^{C} y_{s1}^{(n,c)} \log_{}{p_{x_{s1}}^{(n,c)}},
\end{equation}
where $x_{s1}$ is either $x_{s1}^{2d}$ or $x_{s1}^{3d}$, $N$ is the number of points and $C$ is the number of categories. The segmentation loss of the second source domain is the same. So the final segmentation loss and contrastive loss can be written as:
\begin{equation}
\mathcal{L}_{seg}=\mathcal{L}_{seg}^{s1}+\mathcal{L}_{seg}^{s2},
\end{equation}
\begin{equation}
\mathcal{L}_{ct}=\mathcal{L}_{ct}^{s1}+\mathcal{L}_{ct}^{s2}.
\end{equation}
Finally, we train the model on source domains using Eq. \ref{all}:
\begin{equation}
     \mathcal{L}_{all}=\mathcal{L}_{seg} + \lambda_{ct}  \mathcal{L}_{ct},
   \label{all}
\end{equation}
where $\lambda_{ct}$ is the trade-off to control the importance of the contrastive loss.
\section{Experiments}
\subsection{Datasets and Generalization Settings}
We conduct experiments using three autonomous driving datasets acquired by different LiDAR configurations. (1) A2D2 \cite{a2d2}:  The point clouds are acquired by a Velodyne 16-beam LiDAR. The LiDAR frames are labeled point by point. The data is divided into $\sim$28K training frames and $\sim$2K validation frames. (2) nuScenes \cite{nuscenes}: It contains $\sim$40K LiDAR frames annotated with 3D bounding boxes. Following previous methods \cite{xmuda,dualcross,dscml,sse}, we assign point-wise labels based on the 3D bounding box where points are located. Unlike the A2D2 dataset, it uses a 32-beam LiDAR sensor with different configurations, resulting in a sampling gap from the A2D2 point clouds. We train our model on $\sim$28K frames from 700 training scenes and evaluate on $\sim$6K frames from 150 validation scenes. (3) SemanticKITTI \cite{semantickitti}: Different from A2D2 and nuScenes, it uses a Velodyne 64-beam LiDAR. We use sequences 00-07 and 09-10 for training and evaluate on sequence 08, resulting in $\sim$19K training frames and $\sim$4K frames for evaluation. For all datasets, the LiDAR and RGB camera are synchronized and calibrated. The projections between 2D pixels and 3D points are provided by data prior. We only use the front camera image and the LiDAR points that project into it for simplicity and consistency across datasets. Only 3D annotations are used for 3D semantic segmentation.
 
To comprehensively evaluate the performance of BEV-DG, we design three generalization settings. (1) A,S$\to$N: the network is trained on samples from A2D2 and SemanticKITTI, but tested on samples from nuScenes. (2) A,N$\to$S: we train the network with A2D2 and nuScenes, but test it with SemanticKITTI. (3) N,S$\to$A: the network is trained on nuScenes and SemanticKITTI, but tested on A2D2. We define 5 shared classes between the three datasets: car, truck, bike, person and background. They are all common and safety-critical in self-driving scenes. 

\subsection{Implementation Details}
\noindent{\textbf{Backbone Network. }}
For a fair comparison, we utilize the same backbone network as the previous methods \cite{xmuda,dualcross,dscml,sse}. Concretely, the 2D network is a modified version of U-Net \cite{unet} with a ResNet34 \cite{resnet} encoder, which is pre-trained on ImageNet \cite{imagenet}. For the 3D network, we use SparseConvNet \cite{sparseconvnet} with U-Net \cite{unet} architecture and implement downsampling for six times. Using a voxel size of $5cm$ in SparseConvNet, we ensure only one 3D point exists in each voxel. Our model is trained and evaluated using the PyTorch deep learning framework on a single NVIDIA TITAN RTX GPU with 24GB RAM.

\noindent{\textbf{Parameter Settings. }}
We choose a batch size of $8$ and Adam optimizer with $\beta_1 = 0.9$ and $\beta_2 = 0.999$. We utilize an iteration-based learning schedule where the initial learning rate is $0.001$, and then it is divided by $10$ at $80k$ and $90k$ iterations. Our model is trained for $100k$ iterations on each generalization setting. The $w$ in Eq. \ref{eq:voxel feature} is set to $0.5m$. Furthermore, we set both $\tau$ in Eq. \ref{conloss} and $\lambda_{ct}$ in Eq. \ref{all} to 0.01. The segmentation accuracy is evaluated by mean Intersection over Union (mIoU).
\subsection{Experimental Results}
\begin{table*}[t]
  \centering
      \caption{Quantitative results (mIoU, \%) in different domain generalization settings. The baseline architecture only contains 2D and 3D backbone networks with segmentation heads. The training of it is supervised by segmentation loss of source domains. In `Oracle', we train and test the baseline model on the target domain. `Avg' is the ensembling result obtained by taking the mean of the predicted 2D and 3D probabilities after softmax. The best value is marked in {\color{red}red}, and the second best value is marked in {\color{blue}blue}. }
    \resizebox{0.8\linewidth}{!}{
\begin{tabular}{@{}cllccccccccccccc@{}}
\toprule
\multicolumn{3}{c}{\multirow{2}{*}{Method}} & \multicolumn{3}{c}{A,S$\to$N} &  &  & \multicolumn{3}{c}{A,N$\to$S} &  &  & \multicolumn{3}{c}{N,S$\to$A} \\ \cmidrule(lr){4-6} \cmidrule(lr){9-11} \cmidrule(l){14-16} 
\multicolumn{3}{c}{}                       & 2D      & 3D     & Avg    &  &  & 2D      & 3D     & Avg    &  &  & 2D      & 3D     & Avg    \\ \midrule
\multicolumn{3}{c}{Baseline}               & 48.5    & \color{blue}49.4   & 53.9   &  &  & 32.1    & 51.4   & 44.7   &  &  & 48.0    & 45.0   & 48.8   \\ \midrule
\multicolumn{3}{c}{xMUDA\cite{xmuda}}                  & 49.8    & 49.1   & 55.9   &  &  & 32.8    & 52.0   & 44.9   &  &  & 50.9    & 44.9   & 50.8   \\
\multicolumn{3}{c}{xMUDA Fusion\cite{xmuda}}          & 46.8    & 49.0   & 52.8   &  &  & 32.3    & \color{red}56.5   & 44.7   &  &  & 51.5    & 45.4   & 50.4   \\
\multicolumn{3}{c}{DsCML\cite{dscml}}                  & 48.2       & 47.6      & 52.3      &  &  & 31.6       & 51.3      & 43.8      &  &  & 52.2       & \color{blue}46.1      & 51.7      \\
\multicolumn{3}{c}{SSE-xMUDA\cite{sse}}                  & 44.9       & 48.6      & 53.9      &  &  & \color{blue}36.1       & 52.7      & \color{blue}47.3      &  &  & \color{red}55.3       & 44.8      & \color{blue}52.0      \\
\multicolumn{3}{c}{Dual-Cross\cite{dualcross}}             & \color{blue}50.8    & 48.1   & \color{blue}56.0   &  &  & 32.3    &\color{blue}55.5   & 42.6   &  &  & 53.1    & 41.3   & 50.4   \\
\multicolumn{3}{c}{BEV-DG}                    & \color{red}58.0    & \color{red}59.3   & \color{red}59.0   &  &  & \color{red}47.9    & 54.7   & \color{red}60.2   &  &  & \color{blue}55.0    & \color{red}55.1   & \color{red}56.7   \\ \midrule
\multicolumn{3}{c}{Oracle}                 & 64.8       & 57.9      & 69.0      &  &  & 55.5       & 72.8      & 70.7      &  &  & 81.7       & 53.1      & 82.4      \\ \bottomrule
\end{tabular}}
  \label{tab:results}
\end{table*}
We evaluate the performance of our method on three domain generalization settings, \textit{i.e.}, A,S$\to$N, A,N$\to$S and N,S$\to$A, and compare it with several representative state-of-the-art competitors. These methods share the same 2D and 3D backbone networks as ours. 

\begin{figure}[t]
    \centering
\includegraphics[width=\linewidth]{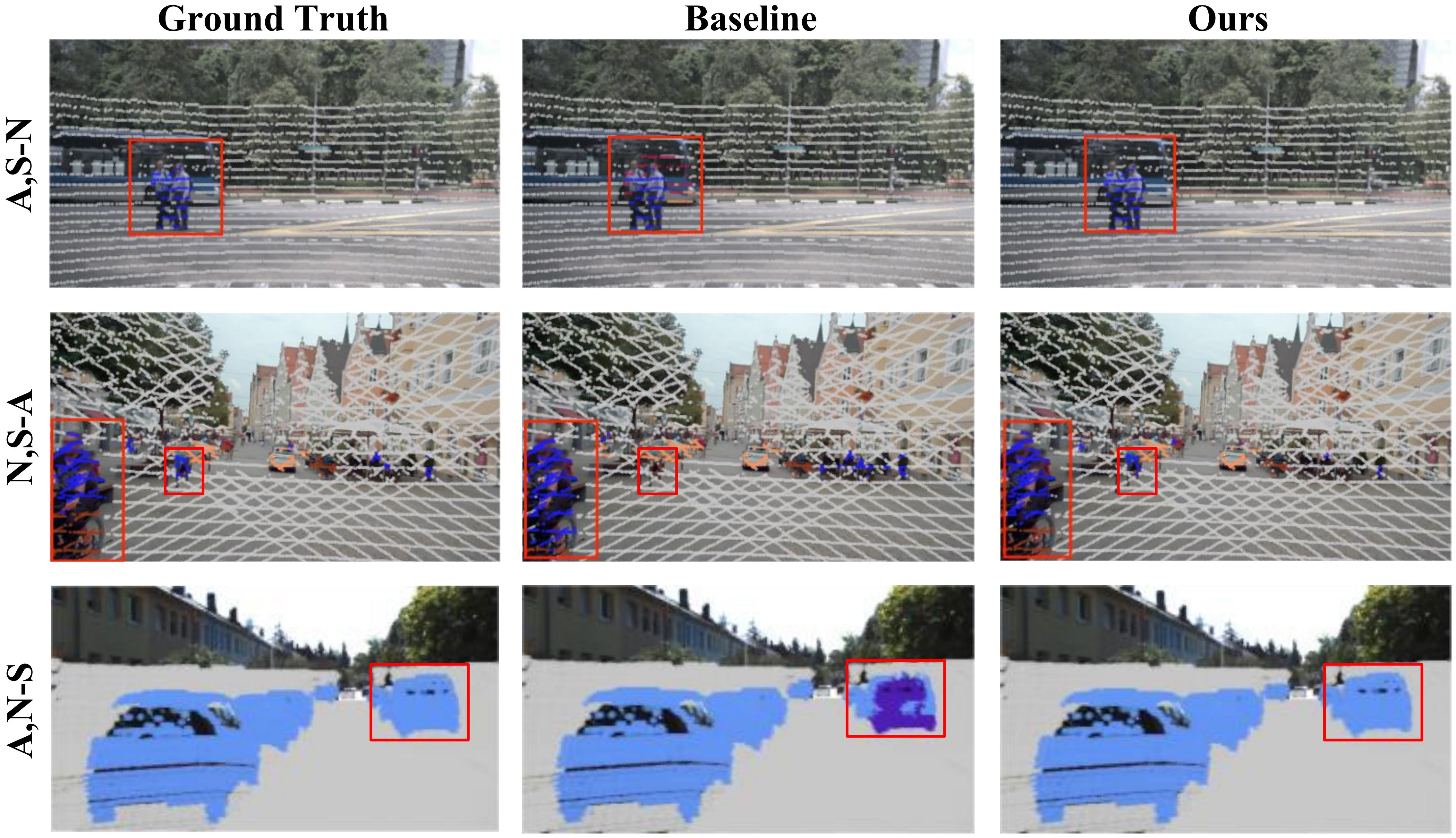}
    \caption{Qualitative results in three domain generalization settings: A,S$\to$N, N,S$\to$A and A,N$\to$S. More qualitative results are shown in the supplementary material.}
    \label{fig:qualitative}
\end{figure}

We visualize some qualitative segmentation examples in Fig. \ref{fig:qualitative} and detail the comparison results for 3D semantic segmentation in Tab. \ref{tab:results}. We can observe that BEV-DG consistently improves results of both 2D and 3D modalities in all three generalization settings compared to competitors. On A,S$\to$N, BEV-DG outperforms the baseline by 9.5\% (2D), 9.9\% (3D) and 5.1\% (Avg). From the first row of Fig. \ref{fig:qualitative}, we can observe that the baseline incorrectly classifies a person as a trunk, but BEV-DG does not. Moreover, compared with the second-best values, BEV-DG outperforms them by 7.2\% (2D), 9.9\% (3D) and 3.0\% (Avg). On A,N$\to$S, BEV-DG outperforms the baseline by 15.8\% (2D), 3.3\% (3D) and 15.5\% (Avg). From the third row of Fig. \ref{fig:qualitative}, we can observe that the baseline incorrectly classifies a car while BEV-DG does not. In addition, BEV-DG outperforms the second-best values by 11.8\% (2D) and 12.9\% (Avg) but is slightly worse on 3D. On N,S$\to$A, BEV-DG outperforms the baseline by 7.0\% (2D), 10.1\% (3D) and 7.9\% (Avg). From the second row of Fig. \ref{fig:qualitative}, we can see that the baseline cannot correctly classify the bike and person, while BEV-DG identifies them precisely. Compared to the second-best values, BEV-DG outperforms them by 9.0\% (3D) and 4.7\% (Avg) but is slightly worse on 2D. These results demonstrate that BEV-DG significantly improves the generalization ability of the model through BEV-based area-to-area fusion and BEV-driven domain contrastive learning.
\begin{table}[t]
  \centering
  	\caption{Ablation study results (mIoU, \%) of two modules. }
   \resizebox{\linewidth}{!}{
    \begin{tabular}{cccc|ccc|ccc}
    \toprule
    & {\multirow{2}[4]{*}{BAF}} & {\multirow{2}[4]{*}{BDCL}} &{} & \multicolumn{3}{c|} {A,S$\to$N} &  \multicolumn{3}{c}{A,N$\to$S} \\
    \cmidrule{5-10}  
      & & &  &  {2D} &  {3D} &  {Avg} &  {2D} &  {3D} &  {Avg} \\
    \midrule
   \#1 & \multicolumn{3}{c|}{Baseline}  &  {48.5} &  {49.4} &  {53.9} &  {32.1} &  {51.4} &  {44.7} \\
    \midrule
    
   \#2 &  { \checkmark} &  {}&{}  &  {56.7} &  {57.9} & {57.8} & {45.5} &  {50.8} & {52.9}  \\
     
   \#3 &  {} &  {\checkmark} &{} &   {50.2} &  {49.6} &  {57.1}  &  {34.2} &  {51.9} &  {49.4} \\
     
   \#4 &  { \checkmark} &  { \checkmark} & {}&  {58.0} &  {59.3} &  {59.0} &  {47.9} & {54.7}  & {60.2} \\
   
    \bottomrule
    \end{tabular}%
    }
  \label{tab:BAFBDCL}%
\end{table}

\subsection{Ablation Studies}
\noindent{\textbf{Effects of BAF and BDCL. }}
To demonstrate the effects of BAF and BDCL, we detail the performance of each module by progressively applying them to the baseline, and the results are reported in Tab. \ref{tab:BAFBDCL}. In BAF, we conduct cross-modal learning area-to-area under bird's-eye view, aiming to alleviate the influence of misalignment. In BDCL, we utilize DVM to generate density-maintained BEV vectors to drive contrastive learning, pushing both 2D and 3D networks to learn domain-irrelevant features jointly. 

From the comparison between \#1 and \#2 in Tab. \ref{tab:BAFBDCL}, we can see that BAF significantly improves the performance of Avg by 3.9\% and 8.2\% on A,S$\to$N and A,N$\to$S, respectively. More importantly, compared to the point-to-point methods \cite{xmuda,dualcross,dscml,sse} in Tab. \ref{tab:results}, BAF consistently outperforms all of them on both 2D and 3D modalities, indicating it effectively mitigates the influence of point-level misalignment. When applying BDCL to baseline, \textit{i.e.}, removing the fusion between $f_{s1}^{bev}$ and $f_{s1}^{2d}/f_{s1}^{3d}$ in BEV-DG, we can observe from \#1 and \#3 that BDCL achieves 3.2\% and 4.7\% improvements on A,S$\to$N and A,N$\to$S in Avg, respectively. Furthermore, we apply both BAF and BDCL to the baseline, and the results are shown in \#4. Our BEV-DG gains 9.5\% (2D), 9.9\% (3D) and 5.1\% (Avg) improvements on A,S$\to$N and 15.8\% (2D), 3.3\% (3D) and 15.5\% (Avg) improvements on A,N$\to$S. These results comprehensively confirm the effects of our BAF and BDCL. 

\noindent{\textbf{Analysis of BAF. }}
There are two crucial steps in the BAF module. One is that we transform initial point-level features into BEV feature maps. The other is that we further fuse the 2D and 3D BEV feature maps for following prediction and contrastive learning. To further demonstrate the effects of BAF, we conduct additional experiments by removing the two steps from BEV-DG respectively, and results are shown in Tab. \ref{tab:BAF}. We first experiment by removing the BEV transformation step. Specifically, in module (a), we directly concatenate initial 2D and 3D features using point-to-pixel projections for semantic segmentation. And then, we generate a global vector by performing max pooling on these concatenated point-level features to drive contrastive learning in module (b). The results are shown in \#1 of Tab. \ref{tab:BAF}. Comparing \#3 and \#1, we can find a sharp drop in the performance, indicating that the BEV feature is the crucial reason why our method works. 
\begin{table}[t]
  \centering
  	\caption{Ablation study results (mIoU, \%) of BEV transformation and fusion of BEV feature maps. }
  { \resizebox{\linewidth}{!}{
  \begin{tabular}{llll|ccc|ccc}
\hline
\multicolumn{4}{c|}{\multirow{2}{*}{Method}}       & \multicolumn{3}{c|}{A,S→N}                                                      & \multicolumn{3}{c}{A,N→S}                                                      \\ \cline{5-10} 
\multicolumn{4}{c|}{}                              & 2D                       & 3D                       & Avg                      & 2D                       & 3D                       & Avg                      \\ \hline
\#1         & \multicolumn{3}{l|}{Ours w/o Trans.}    & \multicolumn{1}{l}{56.1} & \multicolumn{1}{l}{56.2} & \multicolumn{1}{l|}{56.2} & \multicolumn{1}{l}{38.7} & \multicolumn{1}{l}{38.9} & \multicolumn{1}{l}{38.7} \\
\#2         & \multicolumn{3}{l|}{Ours w/o Fusion} & 53.6                     & 48.8                     & 58.3                      & 37.2                        & 50.9                        & 45.3                        \\
\#3         & \multicolumn{3}{l|}{Ours-full}            & 58.0                     & 59.3                     & 59.0                      & 47.9                     & 54.7                     & 60.2                     \\ \hline
\end{tabular}
    }}
  \label{tab:BAF}%
\end{table}%

Next, we experiment by removing the fusion of 2D and 3D BEV feature maps. Specifically, in module (a), we directly fuse initial 2D features with the 2D BEV feature map for semantic segmentation. The 3D branch is the same. Moreover, in module (b), contrastive learning is conducted by using the 2D BEV vector and 3D BEV vector, respectively. The results are shown in \#2 of Tab. \ref{tab:BAF}. Comparing between \#2 and \#3, we can find that ``ours-full'' gains 4.4\% 
 (2D), 10.5\% (3D) and 0.7\% (Avg) improvements on A,S$\to$N and 10.7\% 
 (2D), 3.8\% (3D) and 14.9\% (Avg) improvements on A,N$\to$S, which demonstrates the effectiveness of area-to-area fusion.

\noindent{\textbf{Analysis of BDCL. }}
In BDCL, we propose DVM to generate the density-maintained vector, which can sufficiently embody the domain attributes, to drive contrastive learning. Moreover, we utilize DT to generate the density-transferred vector to form positive pairs. Thus, to further demonstrate the effects of BDCL, we conduct additional experiments by removing the two components from BEV-DG respectively, and results are shown in Tab. \ref{tab:DVMDT}. We first experiment by removing DVM. Specifically, we directly perform max pooling on BEV feature map $f_{s1}^{bev}\in \mathcal{R}^{W\times L\times C}$ to generate a global vector with a size of $1\times C$ to drive contrastive learning. Compared with density-maintained vector $v_{s1}^{bev}$, this vector can not maintain the perception of point cloud density because it is generated by treating different areas of the point cloud equally. The results are shown in \#1 of Tab. \ref{tab:DVMDT}. Comparing between \#1 and \#3, we can find that ``ours-full'' gains 1.9\% (2D), 3.1\% 
 (3D) and 2.3\% (Avg) improvements on A,S$\to$N and 7.5\% (2D), 6.9\% (3D) and 10.2\% (Avg) improvements on A,N$\to$S, which demonstrates the effectiveness of DVM. It also confirms that more domain-related features for contrastive learning can achieve better domain-irrelevant feature learning.
\begin{table}[t]
  \centering
  	\caption{Ablation study results (mIoU, \%) of DVM and DT. }
  { \resizebox{\linewidth}{!}{
  \begin{tabular}{llll|ccc|ccc}
\hline
\multicolumn{4}{c|}{\multirow{2}{*}{Method}}    & \multicolumn{3}{c|}{A,S→N} & \multicolumn{3}{c}{A,N→S} \\ \cline{5-10} 
\multicolumn{4}{c|}{}                           & 2D      & 3D      & Avg    & 2D      & 3D     & Avg    \\ \hline
\#1         & \multicolumn{3}{l|}{Ours w/o DVM} & 56.1    & 56.2    & 56.7   & 40.4    & 47.8   & 50.0   \\
\#2         & \multicolumn{3}{l|}{Ours w/o DT}  & 56.3    & 57.1    & 57.0   & 46.7    & 41.5   & 49.1   \\
\#3         & \multicolumn{3}{l|}{Ours-full}    & 58.0    & 59.3    & 59.0   & 47.9    & 54.7   & 60.2   \\ \hline
\end{tabular}
    }}
  \label{tab:DVMDT}%
\end{table}%

Next, we experiment by removing DT. Specifically, in module (b), we replace density-transferred vector $v_{s1\to s2}^{bev}$ with a copy of $v_{s1}^{bev}$. The results are shown in \#2 of Tab. \ref{tab:DVMDT}. Comparing between \#2 and \#3, we can find that ``ours-full''  achieves 2.0\% and 11.1\% improvements on A,S$\to$N and A,N$\to$S in Avg respectively, which indicates that DT can help the BDCL model domain-irrelevant features by introducing density discrepancy.

\noindent{\textbf{Analysis of Misalignment. }}
To demonstrate the influence of point-to-pixel misalignment, we evaluate two representative point-to-point methods (xMUDA \cite{xmuda} and Dual-Cross \cite{dualcross}) on A,S$\to$N. Furthermore, we compare them with our proposed BAF and BEV-DG, which conduct cross-modal learning area-to-area. We randomly select a fraction of the points in the point cloud and perturb their projections to pixels. The results (Avg) are shown in Fig. \ref{fig:disturb}. We can observe that  BAF and BEV-DG with 20\% misalignment even perform better than xMUDA and Dual-Cross with 5\% misalignment. Moreover, these point-to-point methods degrade more dramatically with increasing misalignment. These results suggest that approaches based on point-to-point cross-modal learning are more sensitive to point-level misalignment. In contrast, with the help of cross-modal learning under bird's-eye view, our BAF and BEV-DG effectively mitigate the influence of misalignment.

\begin{figure}[t]
    \centering
\includegraphics[width=0.9\linewidth]{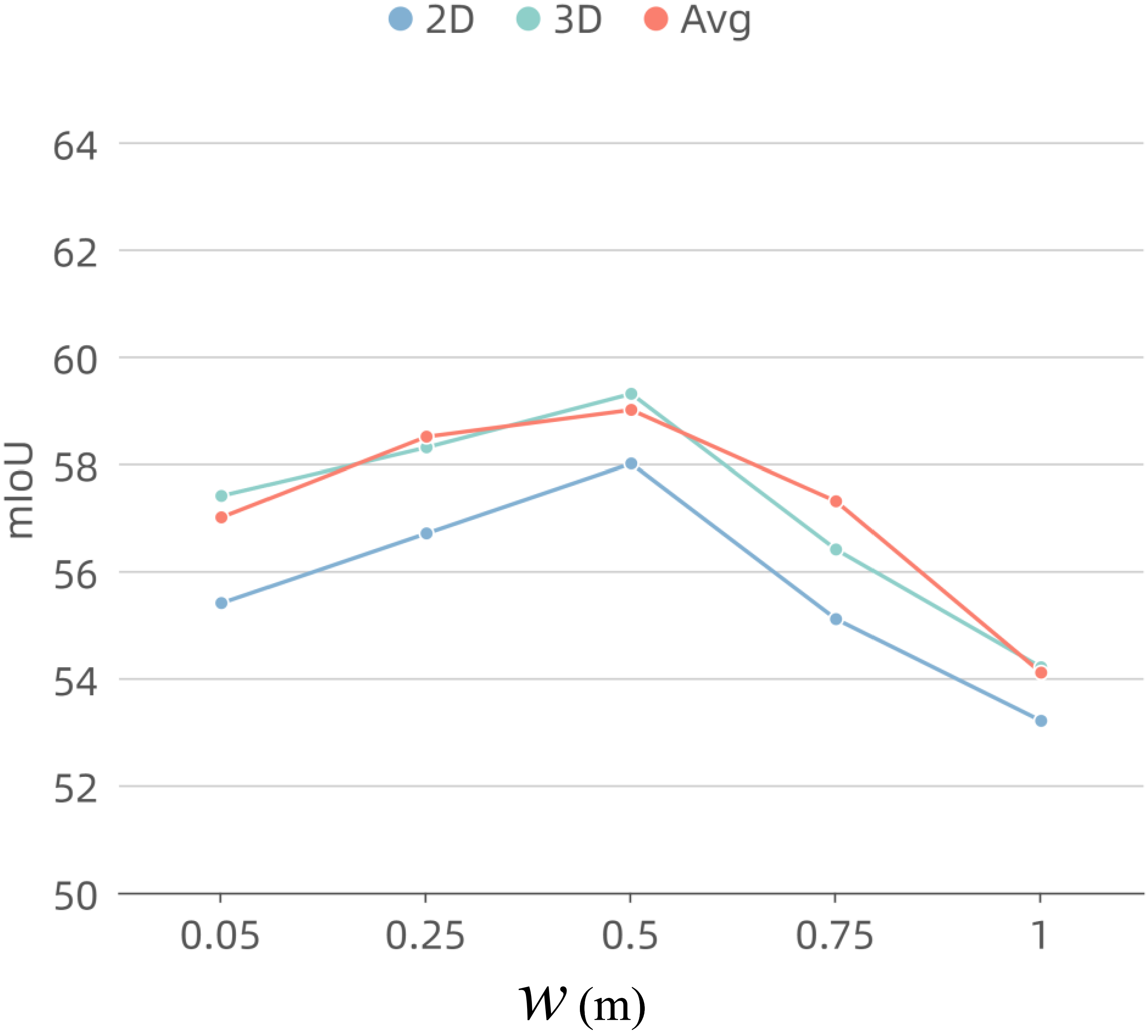}
    \caption{Results of BEV-DG with different $w$.}
    \label{fig:w}
\end{figure}

\begin{figure}[t]
    \centering
\includegraphics[width=0.9\linewidth]{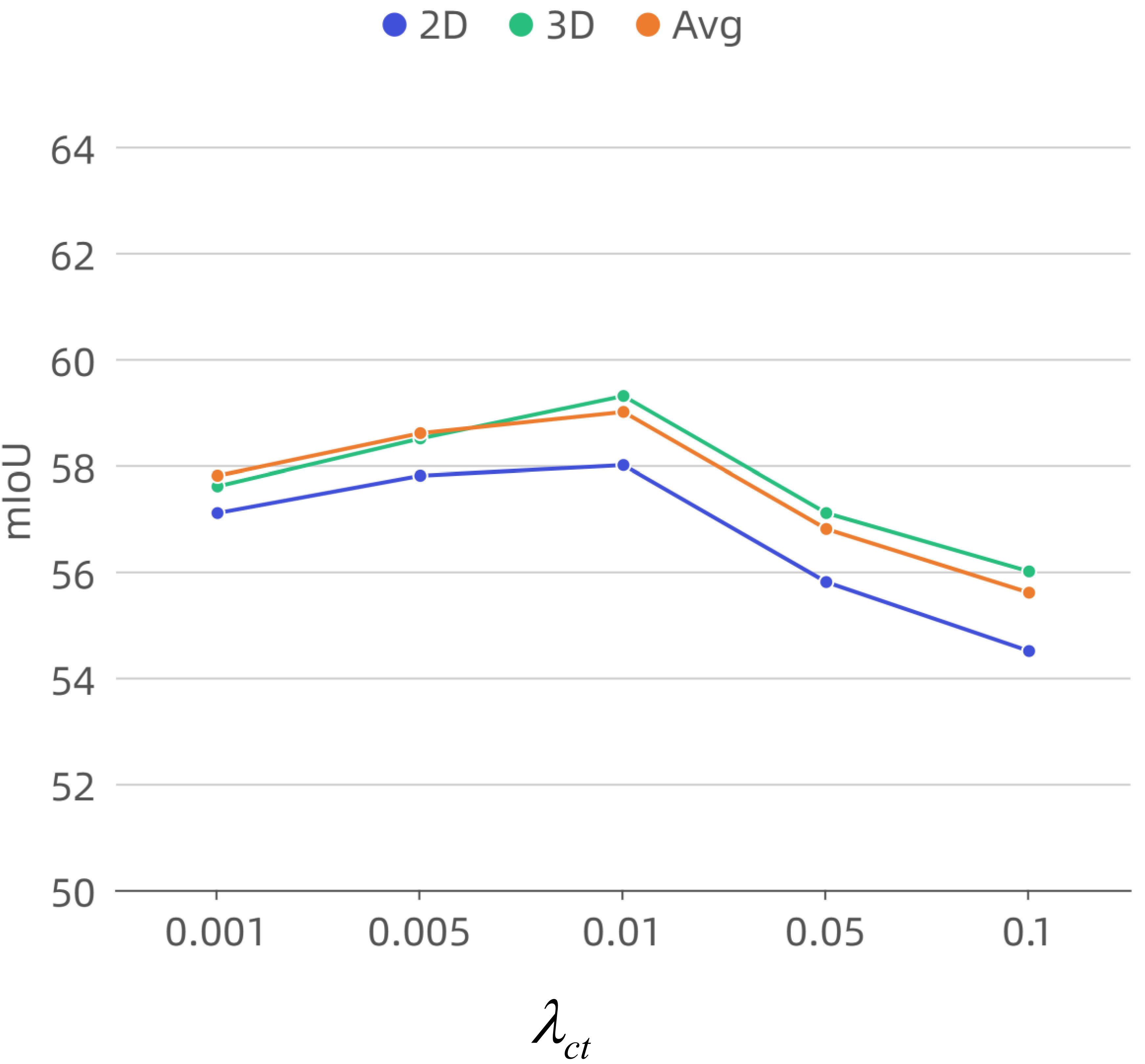}
    \caption{Results of BEV-DG with different $\lambda_{ct}$.}
    \label{fig:lamda}
\end{figure}

\noindent{\textbf{Hyperparameter Sensitivity Analysis. }}
To investigate the impact of two important hyperparameters, \textit{i.e.}, area size $w$ and contrastive loss weight $\lambda_{ct}$, we conduct additional experiments on A,S$\to$N by changing their values in BEV-DG. $w$ is critical to BAF because it determines the size of an area, directly affecting the effectiveness of area-to-area cross-modal learning. We change the value of $w$ between $0.05m$ and $1m$. And the results are shown in Fig. \ref{fig:w}. We can observe that the performance of BEV-DG is best when $w$ is set to $0.5m$. For $\lambda_{ct}$, it controls the importance of contrastive learning loss, which is crucial to BDCL. We change the value of $\lambda_{ct}$ between 0.001 and 0.1. The results are shown in Fig. \ref{fig:lamda}. We can find that the model works best when $\lambda_{ct}$ is 0.01 and our method is not sensitive to $\lambda_{ct}$. Based on the above facts, we set $w$ and $\lambda_{ct}$ to $0.5m$ and $0.01$, respectively.

\begin{figure}[t]
    \centering
\includegraphics[width=\linewidth]{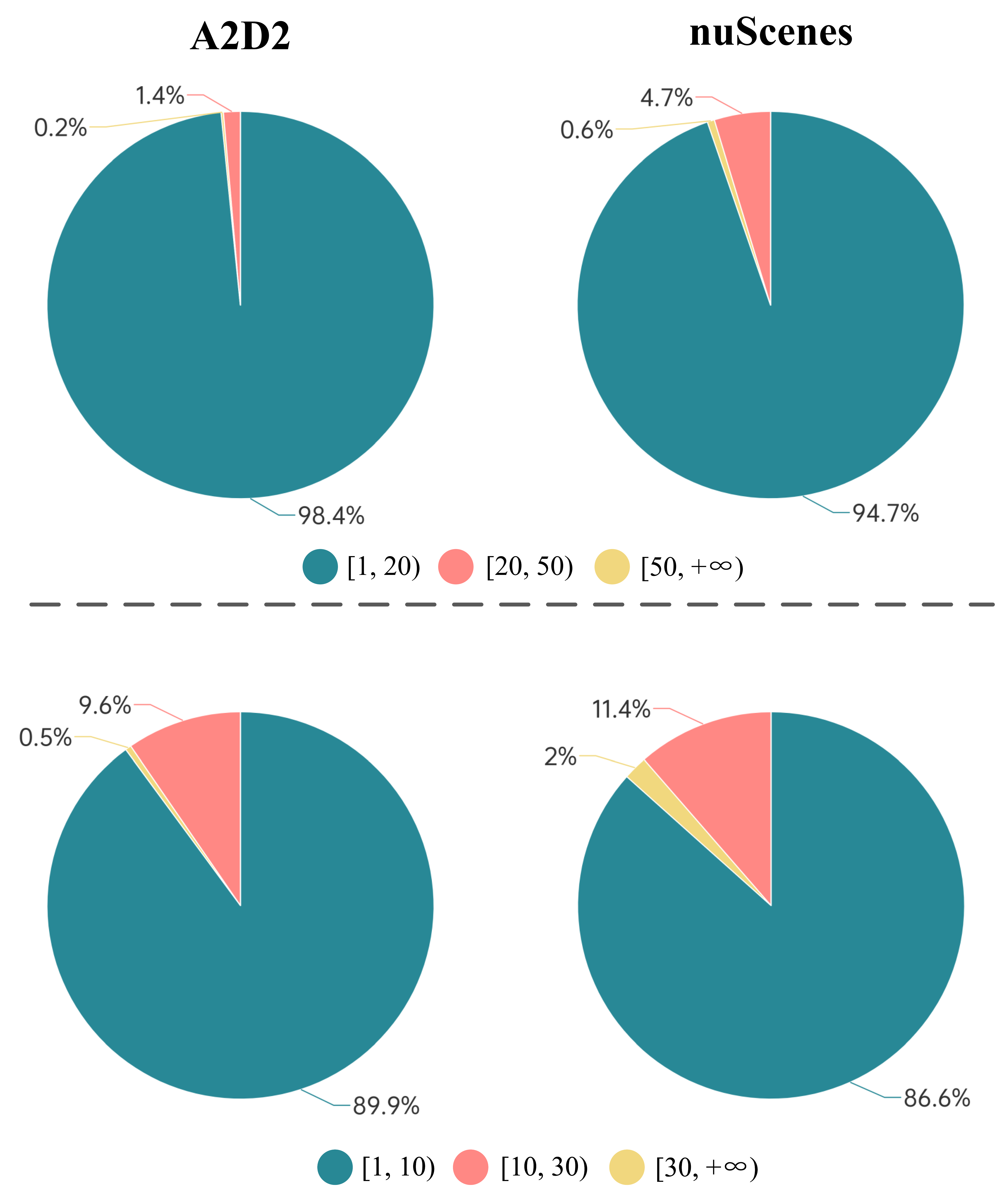}
    \caption{The distribution of areas in BEV space over datasets.}
    \label{fig:newpizza}
\end{figure}

\noindent{\textbf{Area Distribution in the BEV Space. }}
The area distribution in BEV space depends on how to divide areas with different points inside into different types. In BEV-DG, we divide areas into three types, \textit{i.e.},  $[1,10)/[10,50)/[50,+\infty)$, because the distribution pattern generated by this criterion can obviously embody the difference in point cloud density of datasets. To verify the rationality of our criterion for classification, we show the area distribution generated by some other criteria in Fig. \ref{fig:newpizza}. We can observe that these distribution patterns can not obviously embody domain attributes compared to ours, which is shown in Fig. \ref{fig:pizza}.

\section{Conclusion}
This paper proposes cross-modal learning under bird's-eye view for domain generalization of 3D semantic segmentation, aiming to optimize domain-irrelevant representation modeling with the aid of cross-modal learning under bird's-eye view. Specifically, we propose BEV-based area-to-area fusion to achieve cross-modal learning under bird's-eye view, which has a higher fault tolerance for point-level misalignment. Accurate cross-modal learning can more efficiently utilize the complementarity of multi-modality to confront the domain shift. Furthermore, we propose BEV-driven domain contrastive learning to optimize domain-irrelevant representation modeling. With the help of cross-modal learning under bird's-eye-view and density-maintained vector modeling, we generate the BEV vector to drive contrastive learning, pushing the networks to learn domain-irrelevant features jointly. Extensive experimental results on three designed generalization settings demonstrate the superiority of our BEV-DG.

\section*{Acknowledgment}
This work was supported by the National Key Research and Development Program of China No. 2020AAA0108301; National Natural Science Foundation of China under Grant No.62176224; the China Postdoctoral Science Foundation No. 2023M731957.
{\small
\bibliographystyle{ieee_fullname}
\bibliography{egbib}
}
\end{document}